\documentclass[lettersize,journal]{IEEEtran}
\usepackage{amsmath,amsfonts}
\usepackage{algorithmic}
\usepackage{array}
\usepackage[caption=false,font=normalsize,labelfont=sf,textfont=sf]{subfig}
\usepackage{textcomp}
\usepackage{stfloats}
\usepackage{url}
\usepackage{verbatim}
\usepackage{graphicx}
\def\BibTeX{{\rm B\kern-.05em{\sc i\kern-.025em b}\kern-.08em
    T\kern-.1667em\lower.7ex\hbox{E}\kern-.125emX}}
\usepackage{balance}

\usepackage{times}
\usepackage{epsfig}
\usepackage{amsmath}
\usepackage{amssymb}
\usepackage{tabularx}
\usepackage{xcolor}
\usepackage{rotating}
\usepackage{diagbox}
\usepackage{multirow}
\usepackage{bm}

\usepackage{booktabs}
\newcommand{\ours}{LMM-3DP~}

\usepackage[colorlinks,linkcolor=blue]{hyperref}

\makeatletter
\newcommand*\bigcdot{\mathpalette\bigcdot@{.5}}
\newcommand*\bigcdot@[2]{\mathbin{\vcenter{\hbox{\scalebox{#2}{$\m@th#1\bullet$}}}}}
\makeatother

\begin{document}
\title{\textbf{Integrating LMM Planners and 3D Skill Policies for Generalizable Manipulation}}
\author{Yuelei Li$^{\dagger}$, Ge Yan$^{\dagger}$, Annabella Macaluso, Mazeyu Ji, Xueyan Zou, Xiaolong Wang
    \thanks{$^{\dagger}$ Equal Contribution. }
    \thanks{Authors are affiliated with UC San Diego}
}

% \author{IEEE Publication Technology Department 
% \thanks{Manuscript created October, 2020; This work was developed by the IEEE Publication Technology Department. This work is distributed under the \LaTeX \ Project Public License (LPPL) ( http://www.latex-project.org/ ) version 1.3. A copy of the LPPL, version 1.3, is included in the base \LaTeX \ documentation of all distributions of \LaTeX \ released 2003/12/01 or later. The opinions expressed here are entirely that of the author. No warranty is expressed or implied. User assumes all risk.}}

% \markboth{IEEE Transactions on Image Processing, ~Vol.~18, No.~9, Feb~2022}%
% {How to Use the IEEEtran \LaTeX \ Templates}

\maketitle

\begin{abstract}
The recent advancements in visual reasoning capabilities of large multimodal models (LMMs) and the semantic enrichment of 3D feature fields have expanded the horizons of robotic capabilities. These developments hold significant potential for bridging the gap between high-level reasoning from LMMs and low-level control policies utilizing 3D feature fields. In this work, we introduce \ours, a framework that can integrate \textbf{LMM} planners and \textbf{3D} skill \textbf{P}olicies. Our approach consists of three key perspectives: high-level planning, low-level control, and effective integration. \textit{For high-level planning}, \ours supports dynamic scene understanding for environment disturbances, a critic agent with self-feedback, history policy memorization, and reattempts after failures. \textit{For low-level control}, \ours utilizes a semantic-aware 3D feature field for accurate manipulation. \textit{In aligning high-level and low-level control} for robot actions, language embeddings representing the high-level policy are jointly attended with the 3D feature field in the 3D transformer for seamless integration. We extensively evaluate our approach across multiple skills and long-horizon tasks in a real-world kitchen environment. Our results show a significant \textbf{1.45x} success rate increase in low-level control and an approximate \textbf{1.5x} improvement in high-level planning accuracy compared to LLM-based baselines. Demo videos and an overview of \ours are available at \url{https://lmm-3dp-release.github.io}.
\end{abstract}

\section{Introduction}
Building generally capable robots that can perform a wide range of long-horizon tasks in the real world is a long-standing problem. Recent advancements in robotics have been driven by large language models (LLMs) that have shown remarkable capabilities in understanding the real world and common sense reasoning. Some studies leverage LLMs to decompose an abstract task into a sequence of high-level language instructions for planning \cite{ahn2022can, huang2022inner, zeng2022socratic, dasgupta2023collaborating, yao2022react, song2023llm, raman2022planning, singh2023progprompt,wake2023chatgpt, huang2022language}. Despite the significant advancements they have facilitated in various real-world tasks, the current integration of LLMs into robotics presents several major drawbacks. First, LLMs can only process natural language with no visual understanding, making it difficult to comprehend and adapt to dynamic real-world scenarios requiring rich visual information. Additionally, LLM-based planners usually depended on human language feedback to perform long-horizon planning consistently \cite{huang2022language,yao2022react, song2023llm}, which significantly constrains autonomy. However, large multimodal models (LMMs), with multi-sensory inputs, have emerged as a powerful tool to equip robots with strong visual understanding and generalization capabilities across various environments. This allows the robot to adjust language plans according to the environment change. In this paper, we focus on leveraging LMMs to generate language plans based on environment feedback and keep self-improvement in a closed-loop manner.

Existing LLM-based planners typically rely on a predefined set of primitive skills for low-level control \cite{ahn2022can,huang2022inner,chen2023open, codeaspolicy, zeng2022socratic, sayplan}, which is the main bottleneck of large-scale applications to open-world environments. Therefore, the ability to acquire robust low-level skills capable of adapting to the novel environment in a data-efficient manner presents a significant challenge for most LLM-based frameworks. Some recent studies use LLMs to directly output low-level control \cite{yoshida2023text, li2023manipllm}. However, they are only effective in relatively simple manipulation tasks that do not involve rapid high-dimensional control. Due to insufficient 3D understanding, LLMs often fail in complex environments that require comprehending the 3D structure of the scene efficiently. In addition, recent works leverage vision-language models (VLMs) for visual grounding by predicting bounding boxes or keypoints of target objects \cite{huang2023voxposer, ahn2022can}. Despite promising results, they rely on off-the-shelf VLMs which may not be fully optimized for specific, complex tasks in dynamic environments.

To address these challenges, we introduce \ours, an LMM-empowered framework that integrates LMM for self-improved high-level planning and an efficient 3D policy for low-level control. Our framework is designed to satisfy two key requirements: 1) it ensures our LMM agent achieves high autonomy during continuous deployment by decomposing a long-horizon task into high-level plans, calling low-level policy for execution, receiving the environment feedback, and updating language plans accordingly. 2) it allows the low-level policy to learn various skills efficiently with only a few human demonstrations and improve continually.

\begin{figure*}[t!]
\includegraphics[width=1\textwidth]{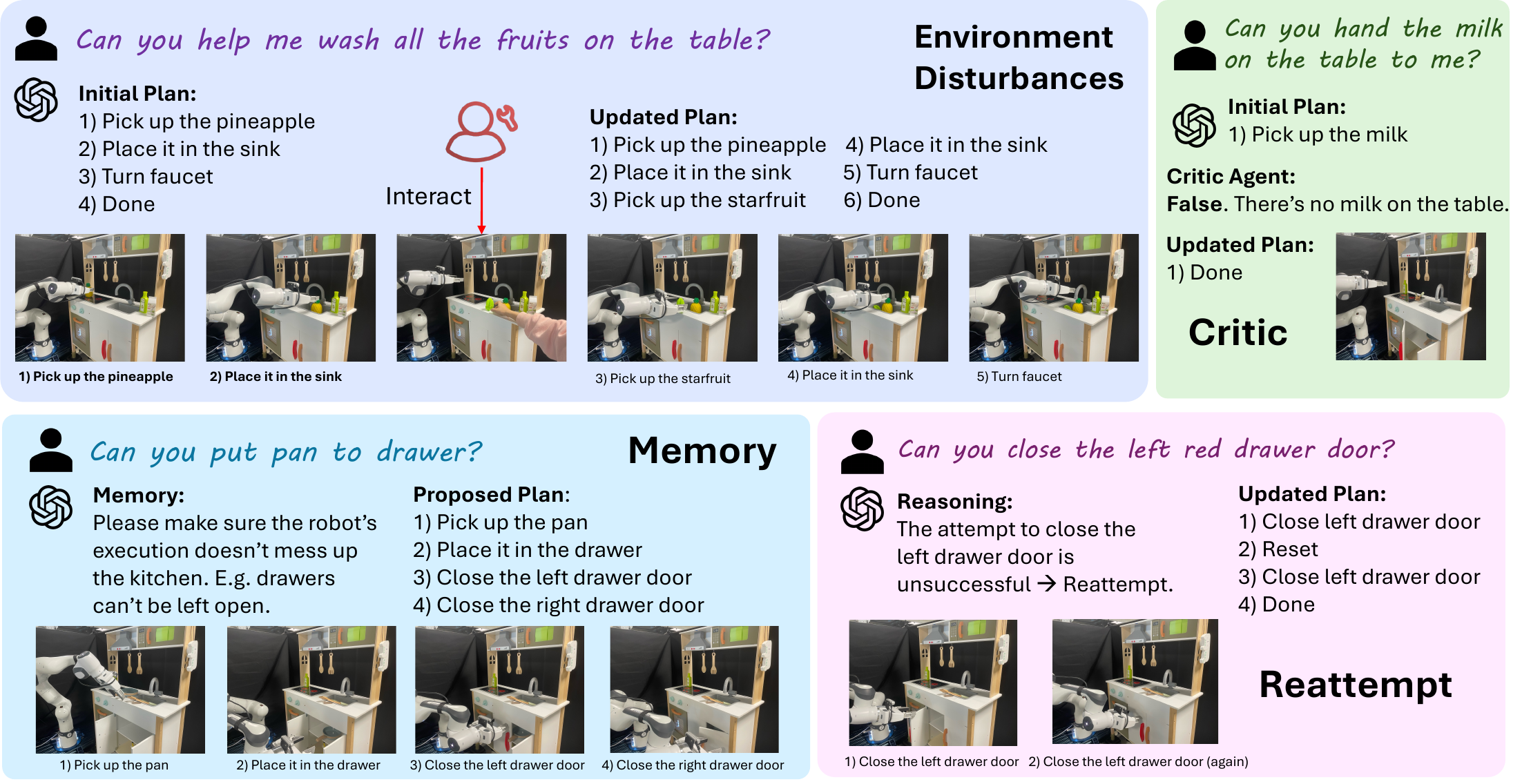}
    \caption{\ours effectively handles environment disturbances, retries if previous attempts fail, and performs accurate reasoning even when human instructions are not aligned with observations.}
    \label{fig:4_examples}
    \vspace{-0.3in}
\end{figure*}

For high-level planning, we introduce three key modules to build an autonomous agent capable of planning a sequence of language instructions: 1) \textbf{Interactive planning with visual feedback.} Incorporating visual feedback within the loop is crucial for enabling an agent to rapidly adapt to dynamic scene changes. In this work, we adopt GPT-4V \cite{achiam2023gpt, gpt4v} as an LMM planner to receive environmental feedback and monitor the ongoing events during execution. 2) \textbf{Self-improvement with memory and critic.} We introduce a critic agent to analyze the plan generated by the LMM planner. It outputs the critique of the planner's decisions and informs whether the plan needs to be updated.  In addition, \ours stores history critique into a memory module and summarizes learned experience for the planner. This approach significantly improves planning accuracy and consistency, especially in challenging long-horizon tasks. 3) \textbf{Life-long learning with a skill library.} Open-ended real-world scenarios usually bring an infinite set of tasks with different skill compositions. The ability to acquire new skills in a data-efficient manner is critical for robots to be generally capable of performing various real-world tasks. Thus, \ours builds a skill library to retrieve different skills required by the LMM planner. When requiring new skills, we adopt an efficient imitation learning policy to grasp such skills with limited human demonstrations. 

More specifically, for precise low-level control, we develop a language-conditioned multi-task 3D policy to learn generalizable skills. To tackle challenging tasks with various object categories and complex environments (\textit{e.g.}, partial occlusion, various geometry shapes, and intricate spatial relationships), it is essential to have a comprehensive semantic and geometry understanding of the scene. Therefore, we first use a vision foundation model to extract 2D semantic features from RGB images, which are then back-projected into 3D space. We then fuse the semantic feature with the geometric point cloud features from a point-based network \cite{qian2022pointnext}. Based on this unified 3D and semantic representation, we train an end-to-end imitation learning policy with a 3D transformer architecture. Our approach is capable of learning various skills efficiently, only requiring a limited number of demonstrations. This facilitates the construction of our ever-growing skill library with robust low-level skills that are reusable and generalizable to novel tasks and environments.

For evaluation, we designed a series of experiments to demonstrate our framework's reliable high-level planning, generalizable low-level control, and exceptional performance in long-horizon tasks. For challenging long horizon tasks, \ours have an average accuracy of \textbf{56.5\%}, while our baseline only has an overall average accuracy of \textbf{7\%} and first step average accuracy of \textbf{50\%} (in a multi-step execution). Additionally, we ablate the design of the critic agent and visual feedback in the loop to delve deeper into the contribution of each component in our framework.

\section{Related Work}
\textbf{LLMs as Task Planners}. Recent advancements in large language models (LLMs) have greatly influenced robotics in various applications. Notable methods typically include using LLMs to generate high-level plans \cite{ahn2022can, huang2022inner, zeng2022socratic, dasgupta2023collaborating, wu2023tidybot}. For example, SayCan \cite{ahn2022can} underscores the extraordinary commonsense reasoning ability of LLMs by generating feasible language plans and adopting an affordance function to weigh the skill's likelihood for execution. Some approaches also leverage LLMs to produce programming code or symbolic API as plan \cite{codeaspolicy, lin2023text2motion, singh2023progprompt, chatgptforrobo, zellers2021piglet, huang2023instruct2act, ahn2024autort}. However, these methods only take natural language instructions as input and lack the ability to perceive the world with multimodal sensory observations. Therefore, they cannot adjust the language plans based on environmental feedback, which strongly limits their performance in dynamic real-world environments. Due to the emergency of LMMs, some studies \cite{hu2023look,huang2024copa, wake2023gpt} leverage GPT-4V \cite{achiam2023gpt} for planning with visual input. However, they only use GPT-4V as a fixed planner without critic and self-improvement while we allow the agent to continue exploring and improving in open-world environments.

\begin{figure}[t!]
\includegraphics[width=0.5\textwidth]{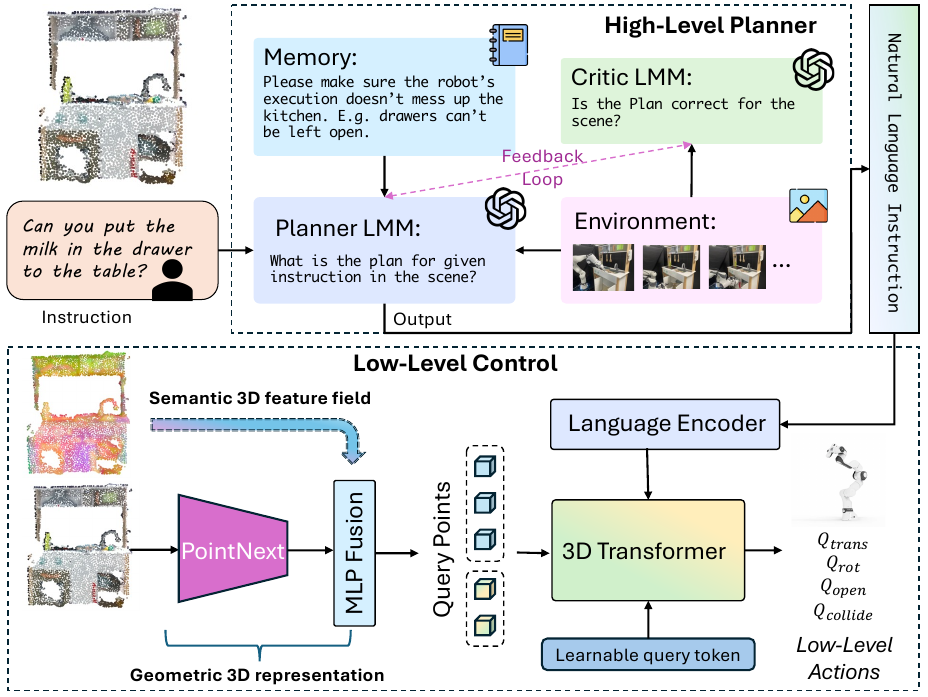}
    \caption{\textbf{\textit{Full Framework Pipeline}}.}
    \label{fig:pipeline}
    \vspace{-0.3in}
\end{figure}

\textbf{Low-Level Robot Primitives.} Despite the significant progress in high-level planning, previous LLM-based language planners \cite{ahn2022can,huang2022inner,chen2023open, codeaspolicy, zeng2022socratic} hold a strong assumption that there exist reliable low-level skills for high-level planners to retrieve, which are usually manually pre-defined set of skills. Some studies \cite{yoshida2023text, codeaspolicy, wang2024llm, cheng2024empowering} use LLMs to output direct low-level control in text, which is impractical to apply to complex real-world tasks requiring high-dimensional control. Some methods \cite{huang2023voxposer, moka, huang2023visual, rana2023sayplan, honerkamp2024language} also leverage vision language models (VLMs) to infer language-grounded affordances and perform motion planning. However, they still lack accurate 3D understanding for challenging environments with diverse geometry shapes and intricate 3D structures. However, \ours addresses this challenge by integrating the high-level planner with a language-conditioned 3D policy that can efficiently learn new skills with a comprehensive 3D understanding of the scene structure.

\textbf{3D Representations for Low-Level Skills.} To learn a visual imitation learning policy for various skills, most previous works \cite{chi2023diffusion,florence2022implicit, ha2023scaling, brohan2022rt, brohan2023rt, padalkar2023open} have been leveraging 2D image-based representation for policy training, while the advantage of 3D representation over 2D images has been increasingly recognized by recent studies \cite{shridhar2023perceiver, ze2023gnfactor, yan2024dnact, chen2023polarnet, zhu2023learning, zhang2023universal}. GNFactor \cite{ze2023gnfactor} and DNAct \cite{yan2024dnact} learn a 3D representation by distilling 2D features from vision foundation models. However, they still require laborious multi-view image collection to train a NeRF \cite{nerf} model, which poses a challenge to large-scale deployment. In this work, we learn a unified 3D and semantic representation by adopting a two-branch architecture with PointNext \cite{qian2022pointnext} and DINO \cite{caron2021emerging} to provide geometry and semantic understanding respectively. Our policy is capable of learning multiple skills with only a few demonstrations.
\section{Method}
In this work, we aim to develop a robust planning framework to generate high-level language plans, along with a generalizable skill-level control policy to follow language plans and execute actions. In this section, we first discuss the design of our self-improved high-level planner, then introduce our language-conditioned skill-level policy (see Figure \ref{fig:pipeline} for the whole pipeline).

\subsection{LMM for High-Level Planning.}
\textbf{Planning with Visual Feedback.} 
In the real world, the optimal plan to execute a task may not be the one initially devised. For instance, you might plan to put vegetables in your favorite blue bowl for dinner, but upon discovering that the blue bowl is unavailable, you use a red bowl instead. Similarly, in robotic planning, the robot must be able to update its plan according to the current situation, which necessitates visual feedback during task execution. We integrate GPT-4V as a planner within the robot's execution loop, enabling it to update the plan after each skill is executed. This design enhances the robot's ability to adapt to dynamic scenes (e.g. when there are environmental disturbances) and reattempt a previous skill if the low-level control fails to execute.

\begin{figure}[t]
    % \vspace{-0.1in}
\includegraphics[width=.5\textwidth]{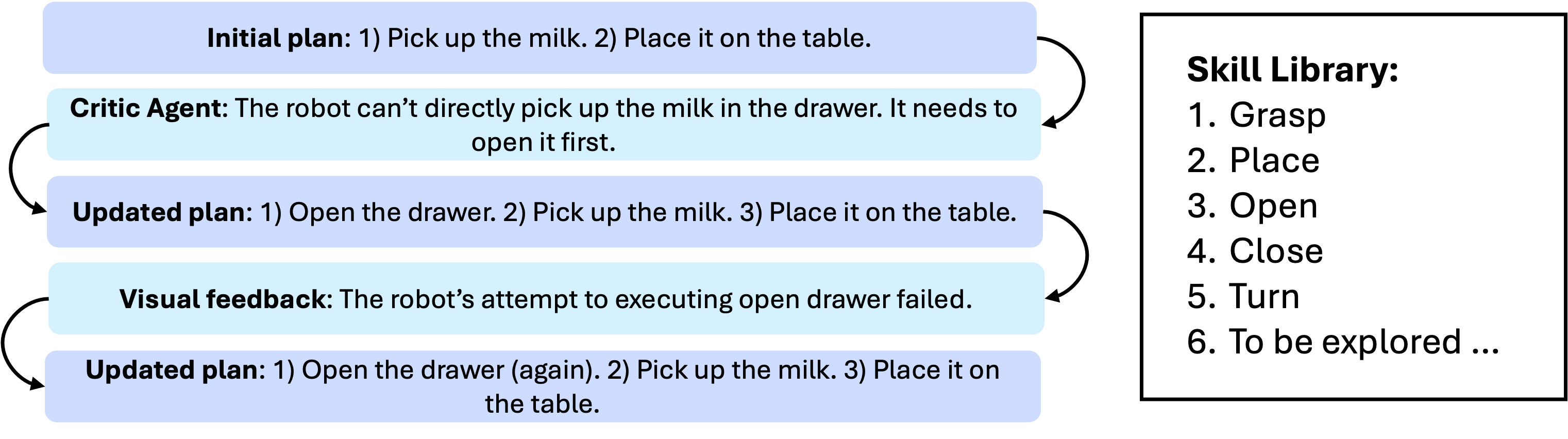}
    \caption{Example of how our planner updates the plan during the robot's execution.}
    \vspace{-0.2in}
    \label{fig:policy}
\end{figure}

\textbf{Critic Agent.} To ensure that the plan generated by the planner is correct and reliable, we introduce an additional critic agent to proactively identify flaws in the generated plan with continuous self-improvement. The critic agent, which only takes visual observation and proposed plan as input (without human instruction), checks whether the next step is feasible in the current situation. If the critic finds that executing the next step will result in an undesirable outcome, its reasoning is input back to the planner, which then proposes a new plan. For instance, the planner's output can be easily skewed by human instructions. This issue persists even with popular prompting techniques \cite{chainofthought}. If the human instruction is to close all the drawers, but some drawers are already closed in the scene, the planner might still generate a plan that involves closing all the drawers. However, the critic can accurately determine that the robot should not close a drawer that is already closed, thereby correcting the planner's mistake.

\textbf{Lifelong Learning.} 
We aim for the planner to improve over time and avoid repeating mistakes by learning from past experience, similar to human learning. However, fine-tuning the planner is computationally expensive. Instead, we employ human critiques of the GPT-4V's output plan and reasoning and then summarize these critiques for in-context learning. These summaries are stored as memory for the planner to reference in the future. Additionally, the planner can propose new skills to the skill library when necessary, then the low-level policy will be updated accordingly with these new skills. For example, in a cooking task, without the click skill, the robot cannot turn on the stove. The planner would identify the click skill as essential for future learning. This approach enables our framework to handle more complex tasks as the skill library expands.

\subsection{Skill learning with 3D Semantic Representation}
Given the language instructions generated by the planner, we train a language-conditioned 3D policy to learn the required low-level skills from human demonstration data. Instead of predicting every continuous action, we extract keyframe actions and convert the skill learning into a keyframe-based action prediction problem. This approach simplifies continuous control and is more sample-efficient for learning a generalizable policy capable of handling novel objects and environments.

\textbf{Vision and language encoder.} To tackle complex real-world environments with various objects and scene structures, we learn a unified 3D and semantic representation by adopting a two-branch architecture: i) Pre-trained with internet-scale data, the vision foundation model has achieved great success in understanding complex scenes with strong zero-shot generalization ability. To leverage these powerful vision foundation models in robotics, we apply a foundation model (e.g., DINO \cite{caron2021emerging}) to extract 2D image features with rich semantics. We then obtain a 3D point feature by back-projecting the 2D feature maps to 3D space. ii) Despite rich semantics from the vision foundation model, it still lacks accurate geometric understanding. Therefore, we adopt a separate branch of a point-based model (e.g., PointNext \cite{qian2022pointnext}) to learn a geometry point feature for better capturing local 3D structures. Subsequently, both semantic and geometry point features are fused by an MLP layer. To incorporate language understanding into our policy,  we use a pre-trained language encoder from CLIP \cite{radford2021learning} to obtain a language embedding.

\begin{table}[t]
\caption{Skill Accuracy.}
\resizebox{0.5\textwidth}{!}{
    \begin{tabular}{c|p{0.7cm}|p{0.7cm}|p{0.7cm}|p{0.7cm}|p{0.7cm}}
        \toprule
        Ours & Grasp & Place & Turn & Open & Close \\
        \midrule
        w/o distractor & 90\% & 65\% & 80\% & 40\% & 100\% \\
        \midrule
        w/ distractor & 56\% & 50\% & 70\% & 40\% & 80\% \\
        \bottomrule
    \end{tabular}
    }
    \label{table:skill_acc}
    \vspace{-0.15in}
\end{table}

\begin{table}
\caption{Skill Comparison.}
\centering
\resizebox{0.42\textwidth}{!}{
    \begin{tabular}{c|c|c|c|c|c}
        \toprule
        \multicolumn{2}{c|}{Ours} & \multicolumn{2}{c|}{OWL-v2} & \multicolumn{2}{c}{Voxposer}\\
        \midrule
        Grasp & Place & Grasp & Place & Grasp & Place \\
        90\% & 65\% & 60\% & 45\% & 60\% & 50\%\\
        \bottomrule
    \end{tabular}
    }
    \vspace{-0.3in}
    \label{table:skill_acc_baseline}
\end{table}

\textbf{Keyframe action prediction.}
Given the fused 3D point feature, language embedding, and robot proprioception, we adopt a 3D transformer architecture to predict the 6-DOF pose of the next best keyframe. Instead of predicting continuous action, we simplify the model prediction into translation $a_\text{trans}\in \mathbf{R}^3$, rotation $a_\text{rot}\in {0,1}^{(360/5)3}$, gripper openness $a_\text{open}\in [0,1]$, and collision avoidance $a_\text{collision}\in [0,1]$. Specifically, we approximate the continuous 3D field via sampling a fixed set of query points in the gripper’s workspace. We do this because, unlike voxel-based methods that discretize the output space and are memory inefficient, the sampling-based approach provides a continuous output space and saves memory during training. We also define a learnable token to attend to the local structures more efficiently.  Both the query points and the learnable token are passed through multiple cross-attention layers with the visual and language features, to obtain the token feature $f_\mathbf{t}$ and query point feature $f_\mathbf{q}$. We then assign a score for each query point by computing the inner product of $f_\mathbf{t}$ and $f_\mathbf{q}$. The next best waypoint $P_i$ is chosen by applying an argmax operation to the score. Inspired by \cite{gervet2023act3d}, we subsequently resample a reduced set of query points around $P_i$ and refine the selection of waypoints among these query points based on previous predictions.
\begin{align}
    \mathcal{L}_{\text{bc}}=&-\lambda_{\text{trans}} \mathbb{E}_{\mathcal{Q}_{\text{trans}}, Y_{\text{trans}}}\left[LS(\mathcal{V}_{\text{trans}}, Y_{\text{trans}}, \alpha)\right] -\\
    &-\lambda_{\text{rot}} \mathbb{E}_{Y_{\text{rot}}}\left[\log \mathcal{V}_{\text{rot}}\right]-\lambda_{\text{open}} \mathbb{E}_{Y_{\text{open}}}\left[\log \mathcal{V}_{\text{open}}\right]-\\
    &-\lambda_{\text{collide}} \mathbb{E}_{Y_{\text{collide}}}\left[\log \mathcal{V}_{\text{collide}}\right]\,\nonumber
\end{align}
where \( \mathcal{V}_i = \operatorname{softmax}(\mathcal{Q}_i) \) for \( \mathcal{Q}_i \in [\mathcal{Q}_{\text{trans}},
\mathcal{Q}_{\text{rot}},
\mathcal{Q}_{\text{open}}, \mathcal{Q}_{\text{collide}}] \). \( Y_i \in [Y_{\text{trans}}, Y_{\text{rot}}, Y_{\text{open}}, Y_{\text{collide}}] \) is the ground truth one-hot encoding. For translation, we calculate cross-entropy loss between a predicted point index $\mathcal{Q}_{\text{trans}}$ and the ground truth $Y_{\text {trans}}$. We apply a label smoothing function \( LS \) to translation loss to prevent overfitting and mitigate label noise in real-world demonstrations.\(\alpha\) is the smoothing parameter.

\section{Experiments}

\textbf{Experiment Setup \& Implementation Details}. We set up a real-world kitchen environment for our experiments, which is more complicated and has more visual features compared to a simple tabletop setting. Our robot is a 7-DoF Franka Emika Panda robot with a 1-DoF deformable gripper. For visual input, we use two Intel RealSense D435 cameras: one provides a front view, and the other is mounted on the gripper. To collect data for our imitation learning-based low-level policy, we use an HTC VIVE controller and base station to track the 6-DOF poses of human hand movement. Then we use SteamVR to map the controller movement to the end effector of the Franka robot. In low-level policy training, we use 100 human demonstrations for one kitchen setting and 200 demonstrations for two kitchen settings (10 demonstrations for each task). We employ the Adam optimizer with a learning rate of $3 \times 10^{-4}$. The training is conducted on one NVIDIA GeForce RTX 3090 with a batch size of 16. We apply color dropout and translation augmentation techniques to improve the model's performance.

\subsection{Main Results}
To perform well on long-horizon tasks, we need to ensure the following: 1) a generalizable low-level policy capable of performing various skills, 2) an adaptable low-level policy that can compose these skills together, and 3) a situation-aware high-level planner with strong reasoning abilities. We systematically evaluate our framework on each of the three criteria individually, then integrate all these components and test our framework's performance on long-horizon tasks (See Figure \ref{fig:4_examples} for qualitative results). If not otherwise stated, each reported accuracy rate is obtained with 10 trials.

\textbf{Low Level Skills}. We train and evaluate our pipeline's performance across five distinct skills: grasping, placing, turning, opening, and closing. Each skill is tested with various objects and task scenarios (Pick is tested 5 times for each of 5 objects, place 5 times for each of 4 locations, and other skills 10 times total). To show the generalization ability of our low-level policy, we report the individual skill accuracy with and without distractors, where the distractors include 1 - 2 extra toys placed in the kitchen to make it more cluttered (see Table \ref{table:skill_acc}).

\begin{table}[t]
\caption{Pick/Place Accuracy. \textcolor{red}{s} means the object has been trained to be placed in the sink. \textcolor{blue}{d} means the object has been trained to be placed in the drawer.}
    \resizebox{0.5\textwidth}{!}{
    \begin{tabular}{ c | c c c c c c }
     \toprule
      location / object & pineapple (\textcolor{red}{s},\textcolor{blue}{d}) & starfruit (\textcolor{red}{s}) & milk & duck (\textcolor{blue}{d}) & pan \\
     \midrule
     sink & 90\% & 90\% & 80\% & 80\% & 50\%\\
     \midrule
     drawer & 90\% & 80 \% & 70\% & 90\% & 40\%\\
     \bottomrule
    \end{tabular}
    }
    \vspace{-0.1in}
    \label{table:pick_place_acc}
\end{table}

\begin{table}[t]
\caption{High-level Planning Comparison.}
\resizebox{0.5\textwidth}{!}{
\begin{tabular}{ l | c c c}
\toprule
Task & SayCan & Voxposer & Ours \\
\midrule
Open both drawer doors.  & 20\% & \textcolor{teal}{90\%} & \textcolor{teal}{90\%} \\
\midrule
Place the gray pan in the drawer, which is closed initially. & 0\% & 50 \% & \textcolor{teal}{80\%}\\
\midrule
Put the fruits (pineapple, starfruit) into the bowl.  & 40\%  & \textcolor{teal}{100\%} & 90\%\\
\midrule
Stack all the bowls on the kitchen table. & 90\% & \textcolor{teal}{100 \%} & \textcolor{teal}{100 \%}\\
\midrule
Place the pineapple in the sink. & \textcolor{teal}{100 \%} & \textcolor{teal}{100 \%} & \textcolor{teal}{100 \%}\\
\bottomrule
\end{tabular}}
\vspace{-0.2in}
\label{table:accuracy_planning}
\end{table}

We use two baselines. First, we use OWL-v2 \cite{owlv2}, an open-vocabulary object detector as an affordance model to output bounding boxes for different objects. This baseline is similar to the approach used in recent works, like \cite{moka}. We also include Voxposer \cite{huang2023voxposer} as a baseline, which is a recent SoTA method on LLM for long-horizon tasks and robot manipulation. Our results demonstrate that our method significantly outperforms the baseline (see Table \ref{table:skill_acc_baseline}). The detector performs poorly with asymmetrical objects, whereas our method learns to grasp these items from human demonstrations efficiently. Also, the detector is highly view-dependent for locating the center of the object of interest, whereas our method performs well as long as the front camera captures the entire scene.

\textbf{Skill Composition.} To successfully compose different skills in sequence, it is essential to demonstrate that these skills are disentangled so that the execution of one skill does not impact the subsequent skills. Our focus is on pick and place operations, as they are highly interrelated. For instance, after training the model on the tasks of putting object A to location B, and putting object C to location D, the model should also be capable of putting C to B. We randomly combine two locations and five objects. Our results in Table \ref{table:pick_place_acc} show that the pick and place skills can be composed together arbitrarily without extra training. For example, though the milk hasn't been directly trained on being placed in the sink and drawer, the place skill still achieves a high accuracy rate on the milk object.

\textbf{High level planning: GPT4V vs LLM.} We compare our high-level planning method with SayCan \cite{ahn2022can}, a widely used method that leverages LLM reasoning with affordance scores to generate robotic plans, and Voxposer \cite{huang2023voxposer}, a recent SoTA method on LLM for long-horizon tasks and robot manipulation. We found that SayCan's method of selecting the next action based on maximum log-likelihood from an action list limits the language model's reasoning ability. This approach makes the model verb-insensitive, fails to understand the semantic meaning of nouns, and is prone to repeating previous actions. Our proposed framework, similar to Voxposer, directly prompts the planning agent in a conversational format rather than a language completion approach, which produces overall better results. Additionally, both Saycan and Voxposer cannot generate plans that consider the state of objects due to the absence of visual input, whereas our method benefits from visual feedback to generate the most reasonable plan. Our results show that our model performs comparably to Voxposer on common kitchen tasks that do not require visual information for reasoning, but demonstrates superior performance when visual information is necessary (see Table \ref{table:accuracy_planning}). In the ``place gray pan in drawer'' task, our method reliably identifies the closed drawer, opens it, and then places the pan inside. In contrast, SayCan and Voxposer frequently neglect to open the drawer.

\begin{table}[t]
\caption{Two kitchen setting experiments.}
\resizebox{0.5\textwidth}{!}{
\begin{tabular}{l|c|c|c|c|c}
\toprule
 & Grasp & Place & Turn & Open & Close \\
\midrule
1st kitchen (2 kitchen checkpoint) & 72\% & 75\% & 70\% & 50\% & 90\% \\
\midrule
2nd kitchen (2 kitchen checkpoint) & 72\% & 60\% & 70\% & 30\% & 80\% \\
\midrule
Overall (2 kitchen checkpoint) & 72\% & 67.5\% & 70\% & 40\% & 85\% \\
\midrule
1st kitchen (1 kitchen checkpoint) & 90\% & 65\% & 80\% & 40\% & 100\% \\
\bottomrule
\end{tabular}
}
\vspace{-0.1in}
\label{table:2kitchen_experiments}
\end{table}

\begin{table}[t]
\caption{5 vs 10 human demonstrations on two kitchens.}
\resizebox{0.5\textwidth}{!}{
\begin{tabular}{c|c|c|c|c|c}
\toprule
 human demonstrations per task & Grasp & Place & Turn & Open & Close \\
\midrule
10 & 72\% & 67.5\% & 70\% & 40\% & 85\% \\
\midrule
5 & 56 \% & 65\% & 80\% & 50\% & 75\%\\
\bottomrule
\end{tabular}
}
\vspace{-0.2in}
\label{table:human_demo}
\end{table}

\begin{table*}[t]
    % \vspace{-0.1in}
    \caption{Long horizon task accuracy. The notation $(\dotsc)$ refers to the accuracy of successfully finishing the first step.}

    \centering
    \begin{tabular}{p{0.2\textwidth} p{0.4\textwidth} c c c c c c}
     \toprule
     Task Description & Actions & Ours & Saycan + Owl-v2 & Voxposer\\
     \midrule
     Put all the fruits in the sink. &
     1) Pick up the pineapple. 2) Place it in the sink. 3) Pick up the starfruit. 4) Place it in the sink. & \textcolor{teal}{60\%} (80\%) & 20\% (60\%) & 20\% (70\%)\\
     \midrule
     Put the duck in the right drawer and close the drawer doors. &
     1) Pick up the duck. 2) Place it in the right drawer. 3) Close the left red drawer door. 4) Close the right orange drawer door. & \textcolor{teal}{40\%} (90\%) & 0\% (20\%) & 0\% (70\%)\\
     \midrule
     Wash the pineapple. &
     1) Pick up the pineapple. 2) Place it in the sink. 3) Turn faucet. & \textcolor{teal}{70\%} (90\%) & 0\% (40 \%) & 10\% (60\%)\\
     \bottomrule
    \end{tabular}
    \vspace{-0.1in}
    \label{table:long_horizon_acc}
\end{table*}

\begin{table*}[t]
    \caption{Ablation on GPT-4V close-loop planning and reflection. ``Ours w/o close-loop \& critic'' means we only use GPT-4V planning once at the beginning without including the Critic Agent and updating the planning in the following steps.}
    \centering
    \begin{tabular}{ c | c c | c c | c c }
        \toprule
        & \multicolumn{2}{c|}{Voxposer} & \multicolumn{2}{c|}{Ours w/o close-loop \& critic} & \multicolumn{2}{c}{Ours} \\
        \midrule
        random noise & turn faucet & close left drawer & turn faucet & close left drawer & turn faucet & close left drawer\\
        & 0\% & 0\% & 40\% & 40\% & \textcolor{teal}{80\%} & \textcolor{teal}{70\%} \\
        \midrule
        environment disturbances & find fruits & close both drawers & find fruits & close both drawers & find fruits & close both drawers \\
        & 0\% & 0\% & 0\% & 0\% & \textcolor{teal}{60\%} & \textcolor{teal}{50\%} \\
        \midrule
        unaligned instruction & pick pan & open drawer & pick pan & open drawer & pick pan & open drawer \\
        & 0\% & 0\% & 10\% & 50\% & \textcolor{teal}{100\%} & \textcolor{teal}{90\%} \\
        \bottomrule
    \end{tabular}
    \vspace{-0.1in}
    \label{table:close_loop}
\end{table*}
\textbf{Long Horizon Tasks}. Combining skill level control, skill composition, and high-level planning, we design 3 long-horizon tasks and test their accuracy rate. Our first baseline includes a high-level planning module from SayCan \cite{ahn2022can} and a 2D object detection control module using OWLv2 \cite{owlv2}, similar to our previous experiments. We also use Voxposer \cite{huang2023voxposer} as a second baseline. The result (see Table \ref{table:long_horizon_acc} and Figure \ref{fig:4_examples}) shows that our method achieves a much higher accuracy rate compared to the baseline methods. We found that Saycan usually fails with incorrect planning; while Voxposer has better planning ability, it mostly fails due to its suboptimal low-level policy, and its inability to re-plan upon failure attempts. Out planning part, with visual feedback and a critic agent, has nearly 100 \% accuracy. Our mistakes mostly stem from low-level policy, which accumulates through each step.

\textbf{Two Kitchen Results.} To further investigate the generalization ability of our low-level policy, we also trained our model jointly on 2 kitchens and reported the accuracy rates of each skill in each kitchen. Because of the more diverse and complicated data in the two kitchen setting, we notice there is an accuracy decrease of about 10\% to 20\% accuracy in each of the skills (see Table \ref{table:2kitchen_experiments}).

\textbf{Human Demonstrations.} We conduct experiments to show that our model scales effectively with the number of human demonstrations provided. Increasing the number of demonstrations from 5 to 10 per task significantly improves the performance of the grasping and placing skills, while other skills can be learned fairly well with 5 demonstrations already (see Table \ref{table:human_demo}).

\subsection{Ablation Studies}
We ablate two of our design choices: visual feedback and the critic agent. The key findings are: 1) visual feedback enables the robot to update its initial plan when there are environmental disturbances, and 2) it allows the robot to reattempt a skill if the previous attempt is unsuccessful. 3) The critic agent is essential when free-form language instructions are unclear or do not align with visual observations.

\textbf{Ablation on Planning with Visual Feedback.}
We investigate the advantages of having GPT-4V's visual feedback in the execution loop through ``random noise'' and ``environment disturbances'' experiments (see Table \ref{table:close_loop} and Figure \ref{fig:4_examples}). In the ``random noise'' experiments, uniform random noise is added to the predicted pose to simulate a flawed low-level policy (-0.05 to 0.05 in x/y, 0 to 0.08 in z for the ``turn faucet'' task, and -0.05 to 0.05 in x/y, -0.03 to 0.03 in z for the "close drawer" task), so the robot needs to retry tasks until successful completion. Our observations indicate that our method can replan effectively with visual feedback in the robot's execution loop while using GPT-4V only for one shot at the beginning and can't replan accordingly. Most errors in our methods stem from the robot arm colliding with the kitchen after the noise is introduced. Our baseline Voxposer, however, struggles to turn the faucet or open the kitchen drawer even without noise, thus failing in all the trials. 

In the ``environment disturbances'' experiments, we modify the scene after the robot's initial execution to determine if the in-the-loop update adjusts the plan according to the novel scene. In the ``find fruits'' task, the robot is required to pick up all the fruits in the scene and place them in the sink. Initially, only a pineapple is visible on the table, but a starfruit appears after the robot's first action. The experiment is successful if our framework updates the original plan and places the starfruit in the sink. In the ``close both drawers'' task, the robot is asked to close both drawers. After the robot closes one drawer, a human closes the other. Success is achieved if the framework updates the original plan to avoid closing the already closed drawer again. Our method can adapt to new scenes with a high accuracy rate (see Table \ref{table:close_loop}). Our baseline Voxposer cannot update its plan once the environment changes, leading to failures in all tests again. 

\textbf{Ablation on Critic Agent}
We find the critic agent in the execution loop useful when there is ``unaligned instruction'': human instruction is not fully aligned with visual observation. In the ``pick pan'' task, the robot is asked to pick up a pan not present in the scene. The experiment is successful if the framework correctly reasons and outputs ``Done'' directly. In the "open drawer'' task, the robot is instructed to open a drawer that is already open. Success is achieved if the framework outputs ``Done'' directly without attempting to open it again. While the planner is easily skewed by human instruction, the critic agent can consistently base its reasoning on visual observation, since it does not take in human instruction as input. This corrects the planner, resulting in accurate plan outputs in subsequent iterations (see Table \ref{table:close_loop} and Figure \ref{fig:4_examples}). Our baseline Voxposer has 0\% in ``unaligned instruction'': it misinterprets the scene information by, for example, searching for a pan that is not present and attempting to open a drawer that is already open, which indicates a lack of understanding of the objects’ states.

\section{Conclusion}
In this work, we propose \ours, a framework that includes LMMs as high-level planners and a language-conditioned 3D policy capable of learning various skills with only a few human demonstrations. Our experiments show that our high-level planning surpasses baselines by \textbf{1.5x} and our low-level control outperforms baselines by \textbf{1.45x}. These designs enable a significantly improved ability to handle environment disturbances and unaligned language instructions, execute various low-level skills in sequence, and recover from failed attempts. Our work's limitations include the need for careful prompt crafting, difficulty with tasks requiring continuous trajectories, and challenges in generalizing skills like picking objects to novel items with limited demonstrations. 
% Please see Appendix \ref{appendix:limitations} for further details on limitations and future work.

\bibliographystyle{IEEEtran}
\bibliography{cite}

\begin{thebibliography}{10}
\providecommand{\url}[1]{#1}
\csname url@samestyle\endcsname
\providecommand{\newblock}{\relax}
\providecommand{\bibinfo}[2]{#2}
\providecommand{\BIBentrySTDinterwordspacing}{\spaceskip=0pt\relax}
\providecommand{\BIBentryALTinterwordstretchfactor}{4}
\providecommand{\BIBentryALTinterwordspacing}{\spaceskip=\fontdimen2\font plus
\BIBentryALTinterwordstretchfactor\fontdimen3\font minus \fontdimen4\font\relax}
\providecommand{\BIBforeignlanguage}[2]{{%
\expandafter\ifx\csname l@#1\endcsname\relax
\typeout{** WARNING: IEEEtran.bst: No hyphenation pattern has been}%
\typeout{** loaded for the language `#1'. Using the pattern for}%
\typeout{** the default language instead.}%
\else
\language=\csname l@#1\endcsname
\fi
#2}}
\providecommand{\BIBdecl}{\relax}
\BIBdecl

\bibitem{ahn2022can}
M.~Ahn, A.~Brohan, N.~Brown, Y.~Chebotar, O.~Cortes, B.~David, C.~Finn, C.~Fu, K.~Gopalakrishnan, K.~Hausman \emph{et~al.}, ``Do as i can, not as i say: Grounding language in robotic affordances,'' \emph{arXiv preprint arXiv:2204.01691}, 2022.

\bibitem{huang2022inner}
W.~Huang, F.~Xia, T.~Xiao, H.~Chan, J.~Liang, P.~Florence, A.~Zeng, J.~Tompson, I.~Mordatch, Y.~Chebotar \emph{et~al.}, ``Inner monologue: Embodied reasoning through planning with language models,'' \emph{arXiv preprint arXiv:2207.05608}, 2022.

\bibitem{zeng2022socratic}
A.~Zeng, M.~Attarian, B.~Ichter, K.~Choromanski, A.~Wong, S.~Welker, F.~Tombari, A.~Purohit, M.~Ryoo, V.~Sindhwani \emph{et~al.}, ``Socratic models: Composing zero-shot multimodal reasoning with language,'' \emph{arXiv preprint arXiv:2204.00598}, 2022.

\bibitem{dasgupta2023collaborating}
I.~Dasgupta, C.~Kaeser-Chen, K.~Marino, A.~Ahuja, S.~Babayan, F.~Hill, and R.~Fergus, ``Collaborating with language models for embodied reasoning,'' \emph{arXiv preprint arXiv:2302.00763}, 2023.

\bibitem{yao2022react}
S.~Yao, J.~Zhao, D.~Yu, N.~Du, I.~Shafran, K.~Narasimhan, and Y.~Cao, ``React: Synergizing reasoning and acting in language models,'' \emph{arXiv preprint arXiv:2210.03629}, 2022.

\bibitem{song2023llm}
C.~H. Song, J.~Wu, C.~Washington, B.~M. Sadler, W.-L. Chao, and Y.~Su, ``Llm-planner: Few-shot grounded planning for embodied agents with large language models,'' in \emph{Proceedings of the IEEE/CVF International Conference on Computer Vision}, 2023, pp. 2998--3009.

\bibitem{raman2022planning}
S.~S. Raman, V.~Cohen, E.~Rosen, I.~Idrees, D.~Paulius, and S.~Tellex, ``Planning with large language models via corrective re-prompting,'' in \emph{NeurIPS 2022 Foundation Models for Decision Making Workshop}, 2022.

\bibitem{singh2023progprompt}
I.~Singh, V.~Blukis, A.~Mousavian, A.~Goyal, D.~Xu, J.~Tremblay, D.~Fox, J.~Thomason, and A.~Garg, ``Progprompt: Generating situated robot task plans using large language models,'' in \emph{2023 IEEE International Conference on Robotics and Automation (ICRA)}.\hskip 1em plus 0.5em minus 0.4em\relax IEEE, 2023, pp. 11\,523--11\,530.

\bibitem{wake2023chatgpt}
N.~Wake, A.~Kanehira, K.~Sasabuchi, J.~Takamatsu, and K.~Ikeuchi, ``Chatgpt empowered long-step robot control in various environments: A case application,'' \emph{IEEE Access}, 2023.

\bibitem{huang2022language}
W.~Huang, P.~Abbeel, D.~Pathak, and I.~Mordatch, ``Language models as zero-shot planners: Extracting actionable knowledge for embodied agents,'' in \emph{International conference on machine learning}.\hskip 1em plus 0.5em minus 0.4em\relax PMLR, 2022, pp. 9118--9147.

\bibitem{chen2023open}
B.~Chen, F.~Xia, B.~Ichter, K.~Rao, K.~Gopalakrishnan, M.~S. Ryoo, A.~Stone, and D.~Kappler, ``Open-vocabulary queryable scene representations for real world planning,'' in \emph{2023 IEEE International Conference on Robotics and Automation (ICRA)}.\hskip 1em plus 0.5em minus 0.4em\relax IEEE, 2023, pp. 11\,509--11\,522.

\bibitem{codeaspolicy}
J.~Liang, W.~Huang, F.~Xia, P.~Xu, K.~Hausman, B.~Ichter, P.~Florence, and A.~Zeng, ``Code as policies: Language model programs for embodied control,'' 2023.

\bibitem{sayplan}
\BIBentryALTinterwordspacing
K.~Rana, J.~Haviland, S.~Garg, J.~Abou-Chakra, I.~Reid, and N.~Suenderhauf, ``Sayplan: Grounding large language models using 3d scene graphs for scalable robot task planning,'' 2023. [Online]. Available: \url{https://arxiv.org/abs/2307.06135}
\BIBentrySTDinterwordspacing

\bibitem{yoshida2023text}
T.~Yoshida, A.~Masumori, and T.~Ikegami, ``From text to motion: Grounding gpt-4 in a humanoid robot" alter3",'' \emph{arXiv preprint arXiv:2312.06571}, 2023.

\bibitem{li2023manipllm}
X.~Li, M.~Zhang, Y.~Geng, H.~Geng, Y.~Long, Y.~Shen, R.~Zhang, J.~Liu, and H.~Dong, ``Manipllm: Embodied multimodal large language model for object-centric robotic manipulation,'' \emph{arXiv preprint arXiv:2312.16217}, 2023.

\bibitem{huang2023voxposer}
W.~Huang, C.~Wang, R.~Zhang, Y.~Li, J.~Wu, and L.~Fei-Fei, ``Voxposer: Composable 3d value maps for robotic manipulation with language models,'' \emph{arXiv preprint arXiv:2307.05973}, 2023.

\bibitem{achiam2023gpt}
J.~Achiam, S.~Adler, S.~Agarwal, L.~Ahmad, I.~Akkaya, F.~L. Aleman, D.~Almeida, J.~Altenschmidt, S.~Altman, S.~Anadkat \emph{et~al.}, ``Gpt-4 technical report,'' \emph{arXiv preprint arXiv:2303.08774}, 2023.

\bibitem{gpt4v}
\BIBentryALTinterwordspacing
OpenAI. (2023) Gpt-4v(ision) system card. [Online]. Available: \url{https://cdn.openai.com/papers/GPTV_System_Card.pdf}
\BIBentrySTDinterwordspacing

\bibitem{qian2022pointnext}
G.~Qian, Y.~Li, H.~Peng, J.~Mai, H.~Hammoud, M.~Elhoseiny, and B.~Ghanem, ``Pointnext: Revisiting pointnet++ with improved training and scaling strategies,'' \emph{Advances in Neural Information Processing Systems}, vol.~35, pp. 23\,192--23\,204, 2022.

\bibitem{wu2023tidybot}
J.~Wu, R.~Antonova, A.~Kan, M.~Lepert, A.~Zeng, S.~Song, J.~Bohg, S.~Rusinkiewicz, and T.~Funkhouser, ``Tidybot: Personalized robot assistance with large language models,'' \emph{Autonomous Robots}, vol.~47, no.~8, pp. 1087--1102, 2023.

\bibitem{lin2023text2motion}
K.~Lin, C.~Agia, T.~Migimatsu, M.~Pavone, and J.~Bohg, ``Text2motion: From natural language instructions to feasible plans,'' \emph{Autonomous Robots}, vol.~47, no.~8, pp. 1345--1365, 2023.

\bibitem{chatgptforrobo}
S.~Vemprala, R.~Bonatti, A.~Bucker, and A.~Kapoor, ``Chatgpt for robotics: Design principles and model abilities,'' 2023.

\bibitem{zellers2021piglet}
R.~Zellers, A.~Holtzman, M.~Peters, R.~Mottaghi, A.~Kembhavi, A.~Farhadi, and Y.~Choi, ``Piglet: Language grounding through neuro-symbolic interaction in a 3d world,'' \emph{arXiv preprint arXiv:2106.00188}, 2021.

\bibitem{huang2023instruct2act}
S.~Huang, Z.~Jiang, H.~Dong, Y.~Qiao, P.~Gao, and H.~Li, ``Instruct2act: Mapping multi-modality instructions to robotic actions with large language model,'' \emph{arXiv preprint arXiv:2305.11176}, 2023.

\bibitem{ahn2024autort}
M.~Ahn, D.~Dwibedi, C.~Finn, M.~G. Arenas, K.~Gopalakrishnan, K.~Hausman, B.~Ichter, A.~Irpan, N.~Joshi, R.~Julian \emph{et~al.}, ``Autort: Embodied foundation models for large scale orchestration of robotic agents,'' \emph{arXiv preprint arXiv:2401.12963}, 2024.

\bibitem{hu2023look}
Y.~Hu, F.~Lin, T.~Zhang, L.~Yi, and Y.~Gao, ``Look before you leap: Unveiling the power of gpt-4v in robotic vision-language planning,'' \emph{arXiv preprint arXiv:2311.17842}, 2023.

\bibitem{huang2024copa}
H.~Huang, F.~Lin, Y.~Hu, S.~Wang, and Y.~Gao, ``Copa: General robotic manipulation through spatial constraints of parts with foundation models,'' \emph{arXiv preprint arXiv:2403.08248}, 2024.

\bibitem{wake2023gpt}
N.~Wake, A.~Kanehira, K.~Sasabuchi, J.~Takamatsu, and K.~Ikeuchi, ``Gpt-4v (ision) for robotics: Multimodal task planning from human demonstration,'' \emph{arXiv preprint arXiv:2311.12015}, 2023.

\bibitem{wang2024llm}
P.~Wang, M.~Robbiani, and Z.~Guo, ``Llm granularity for on-the-fly robot control,'' \emph{arXiv preprint arXiv:2406.14653}, 2024.

\bibitem{cheng2024empowering}
G.~Cheng, C.~Zhang, W.~Cai, L.~Zhao, C.~Sun, and J.~Bian, ``Empowering large language models on robotic manipulation with affordance prompting,'' \emph{arXiv preprint arXiv:2404.11027}, 2024.

\bibitem{moka}
F.~Liu, K.~Fang, P.~Abbeel, and S.~Levine, ``Moka: Open-vocabulary robotic manipulation through mark-based visual prompting,'' 2024.

\bibitem{huang2023visual}
C.~Huang, O.~Mees, A.~Zeng, and W.~Burgard, ``Visual language maps for robot navigation,'' in \emph{2023 IEEE International Conference on Robotics and Automation (ICRA)}.\hskip 1em plus 0.5em minus 0.4em\relax IEEE, 2023, pp. 10\,608--10\,615.

\bibitem{rana2023sayplan}
K.~Rana, J.~Haviland, S.~Garg, J.~Abou-Chakra, I.~D. Reid, and N.~Suenderhauf, ``Sayplan: Grounding large language models using 3d scene graphs for scalable task planning,'' \emph{CoRR}, 2023.

\bibitem{honerkamp2024language}
D.~Honerkamp, M.~B{\"u}chner, F.~Despinoy, T.~Welschehold, and A.~Valada, ``Language-grounded dynamic scene graphs for interactive object search with mobile manipulation,'' \emph{IEEE Robotics and Automation Letters}, 2024.

\bibitem{chi2023diffusion}
C.~Chi, S.~Feng, Y.~Du, Z.~Xu, E.~Cousineau, B.~Burchfiel, and S.~Song, ``Diffusion policy: Visuomotor policy learning via action diffusion,'' \emph{arXiv preprint arXiv:2303.04137}, 2023.

\bibitem{florence2022implicit}
P.~Florence, C.~Lynch, A.~Zeng, O.~A. Ramirez, A.~Wahid, L.~Downs, A.~Wong, J.~Lee, I.~Mordatch, and J.~Tompson, ``Implicit behavioral cloning,'' in \emph{Conference on Robot Learning}.\hskip 1em plus 0.5em minus 0.4em\relax PMLR, 2022, pp. 158--168.

\bibitem{ha2023scaling}
H.~Ha, P.~Florence, and S.~Song, ``Scaling up and distilling down: Language-guided robot skill acquisition,'' in \emph{Conference on Robot Learning}.\hskip 1em plus 0.5em minus 0.4em\relax PMLR, 2023, pp. 3766--3777.

\bibitem{brohan2022rt}
A.~Brohan, N.~Brown, J.~Carbajal, Y.~Chebotar, J.~Dabis, C.~Finn, K.~Gopalakrishnan, K.~Hausman, A.~Herzog, J.~Hsu \emph{et~al.}, ``Rt-1: Robotics transformer for real-world control at scale,'' \emph{arXiv preprint arXiv:2212.06817}, 2022.

\bibitem{brohan2023rt}
A.~Brohan, N.~Brown, J.~Carbajal, Y.~Chebotar, X.~Chen, K.~Choromanski, T.~Ding, D.~Driess, A.~Dubey, C.~Finn \emph{et~al.}, ``Rt-2: Vision-language-action models transfer web knowledge to robotic control,'' \emph{arXiv preprint arXiv:2307.15818}, 2023.

\bibitem{padalkar2023open}
A.~Padalkar, A.~Pooley, A.~Jain, A.~Bewley, A.~Herzog, A.~Irpan, A.~Khazatsky, A.~Rai, A.~Singh, A.~Brohan \emph{et~al.}, ``Open x-embodiment: Robotic learning datasets and rt-x models,'' \emph{arXiv preprint arXiv:2310.08864}, 2023.

\bibitem{shridhar2023perceiver}
M.~Shridhar, L.~Manuelli, and D.~Fox, ``Perceiver-actor: A multi-task transformer for robotic manipulation,'' in \emph{Conference on Robot Learning}.\hskip 1em plus 0.5em minus 0.4em\relax PMLR, 2023, pp. 785--799.

\bibitem{ze2023gnfactor}
Y.~Ze, G.~Yan, Y.-H. Wu, A.~Macaluso, Y.~Ge, J.~Ye, N.~Hansen, L.~E. Li, and X.~Wang, ``Gnfactor: Multi-task real robot learning with generalizable neural feature fields,'' in \emph{Conference on Robot Learning}.\hskip 1em plus 0.5em minus 0.4em\relax PMLR, 2023, pp. 284--301.

\bibitem{yan2024dnact}
G.~Yan, Y.-H. Wu, and X.~Wang, ``Dnact: Diffusion guided multi-task 3d policy learning,'' \emph{arXiv preprint arXiv:2403.04115}, 2024.

\bibitem{chen2023polarnet}
S.~Chen, R.~Garcia, C.~Schmid, and I.~Laptev, ``Polarnet: 3d point clouds for language-guided robotic manipulation,'' \emph{arXiv preprint arXiv:2309.15596}, 2023.

\bibitem{zhu2023learning}
Y.~Zhu, Z.~Jiang, P.~Stone, and Y.~Zhu, ``Learning generalizable manipulation policies with object-centric 3d representations,'' \emph{arXiv preprint arXiv:2310.14386}, 2023.

\bibitem{zhang2023universal}
T.~Zhang, Y.~Hu, H.~Cui, H.~Zhao, and Y.~Gao, ``A universal semantic-geometric representation for robotic manipulation,'' \emph{arXiv preprint arXiv:2306.10474}, 2023.

\bibitem{nerf}
\BIBentryALTinterwordspacing
B.~Mildenhall, P.~P. Srinivasan, M.~Tancik, J.~T. Barron, R.~Ramamoorthi, and R.~Ng, ``Nerf: Representing scenes as neural radiance fields for view synthesis,'' 2020. [Online]. Available: \url{https://arxiv.org/abs/2003.08934}
\BIBentrySTDinterwordspacing

\bibitem{caron2021emerging}
M.~Caron, H.~Touvron, I.~Misra, H.~J\'egou, J.~Mairal, P.~Bojanowski, and A.~Joulin, ``Emerging properties in self-supervised vision transformers,'' in \emph{Proceedings of the International Conference on Computer Vision (ICCV)}, 2021.

\bibitem{chainofthought}
J.~Wei, X.~Wang, D.~Schuurmans, M.~Bosma, B.~Ichter, F.~Xia, E.~Chi, Q.~Le, and D.~Zhou, ``Chain-of-thought prompting elicits reasoning in large language models,'' 2023.

\bibitem{radford2021learning}
A.~Radford, J.~W. Kim, C.~Hallacy, A.~Ramesh, G.~Goh, S.~Agarwal, G.~Sastry, A.~Askell, P.~Mishkin, J.~Clark \emph{et~al.}, ``Learning transferable visual models from natural language supervision,'' in \emph{International conference on machine learning}.\hskip 1em plus 0.5em minus 0.4em\relax PMLR, 2021, pp. 8748--8763.

\bibitem{gervet2023act3d}
T.~Gervet, Z.~Xian, N.~Gkanatsios, and K.~Fragkiadaki, ``Act3d: 3d feature field transformers for multi-task robotic manipulation,'' 2023.

\bibitem{owlv2}
M.~Minderer, A.~Gritsenko, and N.~Houlsby, ``Scaling open-vocabulary object detection,'' \emph{Advances in Neural Information Processing Systems}, vol.~36, 2024.

\end{thebibliography}

\end{document}